\documentclass[authoryear,preprint,review,12pt]{elsarticle}

\usepackage{url}

\journal{arXiv}

\begin{document}

\begin{frontmatter}

\title{Programming by Example and Text-to-Code Translation for Conversational Code Generation}
\author[aiz]{Eli Whitehouse}
\ead{eliw55@gmail.com}
\address[aiz]{AI Zwei, Inc., New York, United States}

\author[aiz]{William Gerard}
\ead{whgerard1@gmail.com}

\author[aiz3]{Yauhen Klimovich}
\ead{yauhenklimovich@gmail.com}
\address[aiz3]{AI Zwei, Inc., Belarus}

\author[aiz2,sym]{Marc Franco-Salvador}
\ead{marc.franco@symanto.com}
\address[aiz2]{AI Zwei Tech, Valencia, Spain}
\address[sym]{Symanto Research, Valencia, Spain}

\newpage

\begin{abstract}
Dialogue systems is an increasingly popular task of natural language processing. 
However, the dialogue paths tend to be deterministic, restricted to the system rails, regardless of the given request or input text.
  Recent advances in program synthesis have led to systems which can synthesize programs from very general search spaces, e.g. Programming by Example, and to systems with very accessible interfaces for writing programs, e.g. text-to-code translation, but have not achieved both of these qualities in the same system.  We propose Modular Programs for Text-guided Hierarchical Synthesis (MPaTHS), a method for integrating Programming by Example and text-to-code systems which offers an accessible natural language interface for synthesizing general programs.  We present a program representation that allows our method to be applied to the problem of task-oriented dialogue.  Finally, we demo MPaTHS using our program representation.
\end{abstract}


\end{frontmatter}

\section{Introduction}

Dialogue systems is an increasingly popular task of Natural Language Processing (NLP). 
However, the dialogue paths tend to be deterministic, restricted to the system rails, regardless of the given request or input text.
Recently, program synthesis has emerged aiming to solve that problem.
A program synthesis system is a tool for constructing complete programs from incomplete or partial descriptions.  Program synthesis systems are defined by their interfaces for accepting descriptions, the strategies they use to search for matching programs, and the space of programs that they search over \citep{dimensions-in-program-synthesis-talk-paper}.  The ideal program synthesis system has an accessible interface that anyone can use, and finds the optimal program for any description from a very general search space.  

Recently, two approaches to program synthesis have gained attention for their use in existing or anticipated mass-market applications: Programming by Example (PbE) \citep{flashfill,flashmeta} and text-to-code translation with large language models \citep{evaluating-large-language-models-trained-on-code}.  PbE systems accept a description of the target program's behavior in the form of input-output examples.  This description provides a flexible characterization of the target program which can be used to find solutions in very general search spaces.  However, it does not constitute an accessible interface for users who cannot readily describe the outcomes of their intended programs.  Text-to-code systems accept a description of the target program's syntax in the form of natural language text---the most accessible interface.  However, because they are based on machine translation, text-to-code systems search over a much less general search space, loosely defined by the set of programs for which corresponding natural language annotations of their source code are available.  

On their own, PbE and text-to-code systems do not offer accessible interfaces for describing general programs.  We suggest that to accomplish this, a program synthesis system must combine the strengths of both approaches, allowing natural language to inform the search space of programs that it explores \citep{dreamcoder-larc}.  We refer to this challenge as Conversational Code Generation (CCG).  In this work, we propose a novel CCG method, Modular Programs for Text-guided Hierarchical Synthesis (MPaTHS) that integrates the functionality of PbE and text-to-code systems.  Finally, we present a program representation that is adapted for use in MPaTHS and demo its usage.

\section{Related Work}

\paragraph{Program Synthesis Systems} 
Recent literature has explored text-to-code translation with large language models as a path to program synthesis \citep{program-synthesis-with-large-language-models,codet5,unified-pretraining-for-program-understanding,cotext,systematic-evaluation-of-llms-of-code}. These models have shown great promise in areas where large quantities of natural language annotated code is available.  Our work is deeply related to the line of work culminating in the DreamCoder system on PbE-driven Domain-Specific Language (DSL) synthesis \citep{dreamcoder-main,dreamcoder-bootstrap,dreamcoder-learning-libraries}.  Wong et al.\ \citep{dreamcoder-laps} investigate a method similar to MPaTHS called Language for Abstraction and Program Search (LAPS), which also achieves an integration between natural language interpretation and PbE-driven DSL synthesis in the DreamCoder system.  LAPS does not fulfill the same role as a text-to-code system, but allows information gleaned from natural language to inform and improve the process of DSL synthesis.  The inductive logic programming community has also addressed issues of program synthesis in the logic programming paradigm \citep{ilp-at-30}.  Shin et al.\ \citep{patois} and Iyer et al.\ \citep{concode} present text-to-code systems that utilize syntactic code idioms to simplify the work of a semantic parser.  Their DSL synthesis step serves the same purpose as DSL synthesis in MPaTHS, though it does not allow for semantic parsing to reciprocally inform the DSL generation process.

\paragraph{Task-Oriented Dialogue Systems}
Pasupat et al.\ \citep{pasupat2019span} and Cheng et al.\ \citep{conversational-semantic-parsing} use neural components to map between hierarchical symbolic representation of conversational state.  Andreas et al.\ \citep{task-oriented-dialogue-as-dataflow-synthesis} uses a graph representation which allows a neural semantic parser to express programmatic changes to conversational state.  Andreas et al.\ \citep{task-oriented-dialogue-as-dataflow-synthesis} provide a fixed set of metacomputation operators with which to express these operations.

\section{MPaTHS}
\begin{figure}[!t] 
  \begin{center}
  \includegraphics[width=0.8\linewidth]{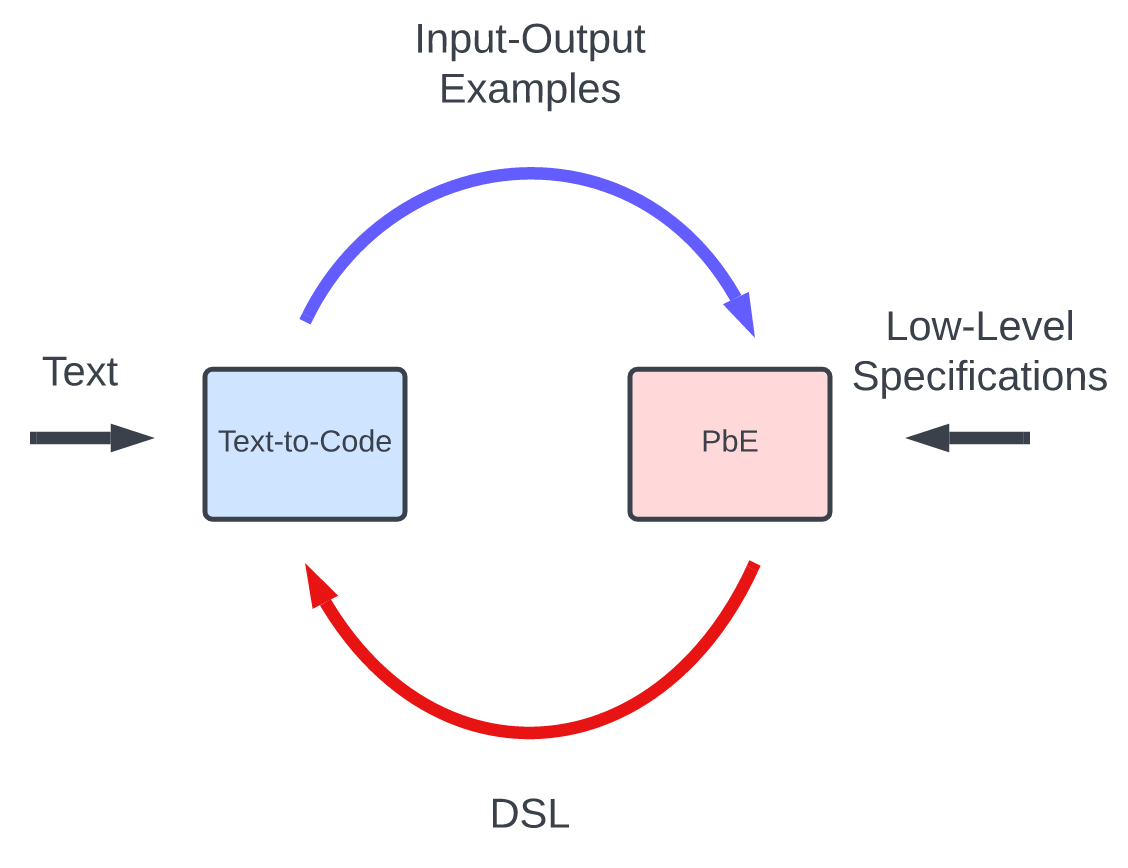}
  \caption{Chart depicting the flow of information between a text-to-code and PbE system in MPaTHS.}
  \label{fig:flow}
  \end{center}
\end{figure}

MPaTHS is a method for utilizing PbE and text-to-code systems synergistically.  When successfully applied, MPaTHS creates a cycle in which a text-to-code system serves as a source of input-output examples, and a PbE system serves as a source of high-level programming operations. Figure~\ref{fig:flow} illustrates this flow.

The main challenge of applying MPaTHS successfully is the design of a program representation that allows this process to proceed while still respecting the input of the domain-modeler who utilizes the resulting system.  Our program representation attempts to balance the expressiveness of domain operations with other requirements on conversational states that hold across domains.  In particular, we aim to make domain operations open-ended and easy to translate from source code, and to make the planning of behavior available by default in all domains.

\section{Program Representation}

Our program representation consists of two layers of specification.  While domain modelers specify the operations of their domain, we define another domain-independent Application Programming Interface (API) of operations for constructing invocations of those operations in the form of graphs. Synthesized programs invoke domain operations by constructing the corresponding graph with this domain-independent API.  

Our graphs generalize on abstract syntax tree representations of programs to allow planning not only within the domain through conditional domain operations, but also at the level of graph structure.  This is achieved by allowing graph modifications which do not have to have an immediate effect of invoking domain operations, but perform preparatory work needed to invoke them later. 

\subsection{Domain-Specific Structure}

Our representation of domain operations is based on the situation calculus of Reiter \citep{reiter-situation-calculus}.  Domain operations have the following components: (a) a procedure name; (b) a list of typed input and output arguments; (c) a precondition stating when the procedure may be executed; (d) a post-condition stating the facts that are established and falsified when the procedure is performed.  While the types of arguments can be utilized to express the same constraints as preconditions and post-conditions, this design allows more flexibility when translating source code: types and procedure names can be directly mirrored from code, while additional properties can be enforced with auxiliary predicates.  

\subsection{Domain-Independent Structure}

Our graphical program representation consists of several categories of structure.  Most of these are used to represent fragments of valid \textit{domain programs}: lists of partial or complete (potentially conditional) domain operations.  We refer to these categories of structure as syntactic structure.  

The syntactic structure of a graph includes an executed path, unexecutable syntax trees, and unexecuted syntax trees.  The executed path represents the program region of invoked domain operations.  Graph construction operations which adjoin the syntax of a domain operation to the executed path or complete the syntax of a domain operation along the executed path have the meaning of invoking these operations. Unexecutable trees are inactive branches of conditional statements which have been evaluated, and play no role in graph construction operations.  Unexecuted trees represent program regions of uninvoked domain operations.  They can be constructed with no immediate effects within the domain, and thus allow for planning (behavior with delayed effects) at the level of graph construction operations. 

Critical to this functionality is extra-syntactic structure, which connects unexecuted trees to executed trees.  Additionally, extra-syntactic structure allows for relationships to be created between domain operations and information about them to be stored causing change within the domain itself. 

\subsection{Example of an Executed Path}

\begin{figure}[!t]
  \begin{center}
  \includegraphics[width=1.0\linewidth]{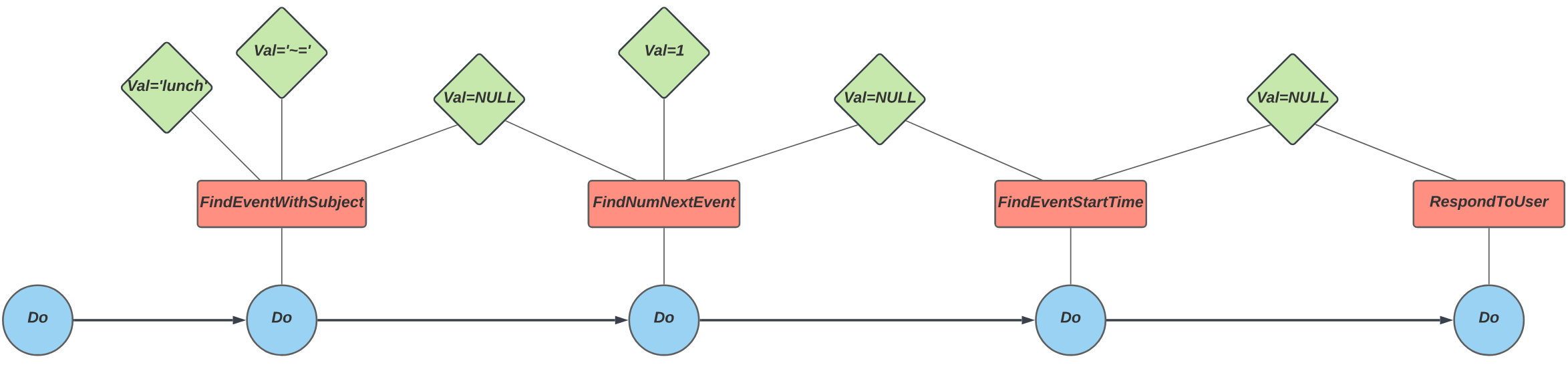}
  \caption{A ground-truth graph for the utterance ``Can you see when my next meeting is involving a lunch?"}
  \label{fig:path}
  \end{center}
\end{figure}

Figure~\ref{fig:path} depicts the executed path of a graph translated from the Lispress graph program representation of the SMCalFlow dataset \citep{task-oriented-dialogue-as-dataflow-synthesis}. Nodes labeled ``Do" act as a hub for the multiple pieces of information that characterize a single domain operation.  The graph program that generated this graph is included in our supplementary materials (see section \ref{sec:suppmat}). In the conference session, we will conduct a live demo of MPaTHS.

\section{Conclusion}

Dialogue systems is a key NLP task.
An important issue of those systems is that the dialogue paths tend to be deterministic, restricted to the system rails, regardless of the given request or input text.
Aiming to solve that issue, our method, MPaTHS, integrates a text-to-code system as a source of input-output examples with a PbE system as the source of a DSL to produce a conversational code generation system. Our program representation allows the underlying domain of application to be specified in the presence of DSL synthesis.  We present this program representation, illustrate it with an example from the SMCalFlow dataset \citep{task-oriented-dialogue-as-dataflow-synthesis}, and demo its usage. In future work, we aim to continue improving on the integration of MPaTHS' modules and extending its capabilities as a task-oriented dialogue system.

\section*{Acknowledgements}

To all present and former members of AI Zwei, whose passion and dedication made this research work possible.

The work from Marc Franco has been partially funded by the Pro$^2$Haters - Proactive Profiling of Hate Speech Spreaders (CDTi IDI-20210776), the XAI-DisInfodemics: eXplainable AI for disinformation and conspiracy detection during infodemics (MICIN PLEC2021-007681), the OBULEX - \textit{OBservatorio del Uso de Lenguage sEXista en la red} (IVACE IMINOD/2022/106), and the ANDHI - ANomalous Diffusion of Harmful Information (CPP2021-008994) R\&D grants.

\bibliographystyle{elsarticle-harv}
\bibliography{references}

\appendix

\section{Supplementary Material} \label{sec:suppmat}
The repository at \url{https://github.com/eliw-aiz/mpaths-FIRE-supplement} contains additional information about our program representation, including a domain-independent DSL and a sample graph program. 

\end{document}